\documentclass[11pt,a4paper]{article}
\usepackage[utf8]{inputenc}
\usepackage[hyperref]{emnlp2018}
\usepackage{times}
\usepackage{latexsym}
\usepackage{amsfonts}       
\usepackage{amsmath}
\usepackage{amssymb}
\usepackage{graphicx}
\usepackage{enumitem}
\usepackage{array}
\usepackage{breakcites}
\usepackage{multirow}

\newcolumntype{L}[1]{>{\arraybackslash}p{#1}}

\usepackage{url}

\aclfinalcopy 

\usepackage[normalem]{ulem}
\definecolor{darkgreen}{rgb}{0.0, 0.5, 0.0}

\def\VRdel#1{\bgroup\markoverwith{\textcolor{magenta}{\rule[0.5ex]{2pt}{1pt}}}\ULon{#1}}

\def\ODdel#1{\bgroup\markoverwith{\textcolor{darkgreen}{\rule[0.5ex]{2pt}{1pt}}}\ULon{#1}}

\def\XXdel#1{\bgroup\markoverwith{\textcolor{orange}{\rule[0.5ex]{2pt}{1pt}}}\ULon{#1}}

\def\YKdel#1{\bgroup\markoverwith{\textcolor{blue}{\rule[0.5ex]{2pt}{1pt}}}\ULon{#1}}

\newcommand*\diff{\mathop{}\!\mathrm{d}} 

\title{Better Conversations by Modeling, Filtering, and Optimizing for Coherence and Diversity}

\author{Xinnuo Xu, Ondřej Dušek, Ioannis Konstas and Verena Rieser \\
  The Interaction Lab, School of Mathematical and Computer Sciences \\
  Heriot-Watt University, Edinburgh \\
  {\tt xx6, o.dusek, i.konstas, v.t.rieser@hw.ac.uk} \\}

\date{}

\makeatletter
\let\thetitle\@title
\makeatother

\begin{document}
\maketitle
\begin{abstract}
We present three enhancements to existing encoder-decoder models for open-domain conversational agents, aimed at effectively modeling coherence and promoting output diversity: (1) We introduce a measure of coherence as the GloVe embedding similarity between the dialogue context and the generated response,
(2) we filter our training corpora based on the measure of coherence to obtain topically coherent and lexically diverse context-response pairs, (3) we then train a response generator using a conditional variational autoencoder model that incorporates the measure of coherence as a latent variable and uses a context gate to guarantee topical consistency with the context and promote lexical diversity. 
Experiments on the OpenSubtitles corpus show a substantial improvement over competitive neural models in terms of BLEU score as well as metrics of coherence and diversity.
\end{abstract}

\section{Introduction}

End-to-end neural response generation methods are promising for developing open domain dialogue systems as they allow to learn from very large unlabeled datasets \cite{Shang:2015,sordoni:2015,Vinyals:2015}. However, these models have also been shown to generate generic, uninformative, and non-coherent replies (e.g., \textit{``I don't know.''} in Figure~\ref{fig:running_example}), mainly due to the fact that neural systems tend to settle for the most frequent options, thus penalizing length and favoring high-frequency word sequences \cite{sountsov_length_2016, wei_why_2017}.

To address these problems, \newcite{JiweiLi:diversity2015} and \newcite{li2017learning} attempt to promote diversity by improving the objective function, but do not model diversity explicitly. \newcite{Serban2017AHL} focus on model structure without any upgrades to the objective function.
Other works control the style of the output by leveraging external resources (\newcite{hu2017toward}: sentiment classifier, time annotation; \newcite{zhao2017learning}: dialogue acts) or focus on well-structured input such as paragraphs \cite{li2016neural}.


This paper extends previous attempts to model diversity and coherence by enhancing all three aspects of the learning process: the data, the model, and the objective function. While previous research has addressed these aspects individually, this paper is the first to address all three in a unified framework. 
Instead of using existing linguistic knowledge or labeled datasets, we aim to control for coherence by learning directly from data, using a fully unsupervised approach.
This is also the first work encoding and evaluating coherence explicitly in the dialogue generation task, as opposed to using diversity, style, or other properties of responses as a proxy.



\begin{figure}
\small
\begin{tabular}{|>{\hspace{-1mm}}l@{~}|>{\hspace{-1mm}}l|}\hline
{\bf Conversational history} & {\bf Response} \\ \hline\hline
A: You stay out of this. & B-Coh: Well, I got \uline{water}. \\ 
B: So you want \uline{water}, huh? & B-Incoh: I don't know. \\ 
A: That's right. & \\\hline\hline
A: Where do we start? & B-Coh: Specifically the \uline{stove}.\hspace{-2mm} \\
B: \uline{Kitchen}. & B-Incoh: Let's go for a walk. \\
A: Definitely the \uline{kitchen}. & \\\hline
\end{tabular}
\caption{Examples of conversational history (left) with two alternative responses to follow it (right): (B-Coh) a more coherent, topical utterance, and (B-Incoh) a generic, inconsistent response.}
\label{fig:running_example}
\end{figure}
In this work, given a dialogue history, we regard as a coherent response an utterance that is thematically correlated and naturally continuing from the previous turns, as well as lexically diverse. For example, in Figure~\ref{fig:running_example} the response \textit{``Specifically the stove.''}\ is a very natural and coherent response, elaborating on the topic of \textit{kitchen} introduced in the previous two utterances and containing rich thematic words, whereas the response \textit{``Let's go for a walk.''}\ is unrelated and uninteresting.

In order to obtain coherent responses, we present three generic enhancements to existing encoder-decoder (E-D) models: 

\begin{enumerate}[itemsep=2pt,leftmargin=12pt]
\item We define a \textit{measure of coherence} simply as the averaged word embedding similarity between the words of the context and the response computed using GloVe vectors \cite{pennington2014glove}.


\item We filter a corpus of conversations based on our measure of coherence, which leaves us with context-response pairs that are both topically coherent and lexically diverse.

\item We train an E-D generator recast as a conditional Variational Autoencoder \cite[cVAE;][]{zhao2017learning} model that incorporates two latent variables, one for encoding the context and another for conditioning on the measure of coherence, trained jointly as in  \newcite{hu2017toward}. 
We then decode using a context gate \cite{tu2016context} to control the generation of words that directly relate to the most topical words of the context and promote coherence.
\end{enumerate}

Experiments on the OpenSubtitles \cite{lison_automatic_2016} corpus demonstrate the effectiveness of the overall approach. Our models achieve a substantial improvement over competitive neural models.
We provide an 
ablation 
analysis, quantifying the contributions that come from effective modeling of coherence into our models.
All our experimental code is freely available on GitHub.\footnote{\url{https://github.com/XinnuoXu/CVAE_Dial}}

\section{Coherence-based Dialogue Generation}\label{sec:task}


Our model aims to generate responses given a dialogue context, incorporating measures of coherence estimated purely from the training data. We propose the following enhancements to the attention-based E-D architecture \cite{bahdanau2014neural,luong2015effective}: 
\begin{itemize}[itemsep=2pt,leftmargin=12pt]
  \item We introduce a stochastic latent variable $\mathbf{z}$ conditioned on previous dialogue context to store the global information about the conversation \cite{bowman2015generating, chung2015recurrent, li2016neural, hu2017toward}.
  \item We force the model to condition on the measure of coherence explicitly by encoding a 
  latent variable (code) $c$ learned from data.

\item We incorporate a context gate \cite{tu2016context} that dynamically controls the ratio at which the generated words in the response derive directly from the coherence-enhanced dialogue context or the previously generated parts of the response.
\end{itemize}

In the rest of this section, we introduce the measure of coherence (Section~\ref{SubSec:coherence_def}), we present an overview of our model (Section~\ref{SubSec:overview}), and finally describe the model in detail (Sections~\ref{SubSec:generator}--\ref{SubSec:testing}).

\subsection{Measure of Dialogue Coherence} \label{SubSec:coherence_def}

Semantic vector space models of language represent each word with a real-valued word embedding vector \cite{pennington2014glove}. By simply taking a weighted average of all its word embeddings, a whole sentence can be mapped into the semantic vector space. We define the coherence of a dialogue as the average distance between semantic vectors of preceding dialogue context and its response.

Let $\mathbf{x} = \left \{ x_{1}, \dots x_{j}, \dots x_{J} \right \}$ represent a dialogue context and $\mathbf{y} = \left \{ y_{1}, \dots y_{i}, \dots y_{I} \right \}$ a response. $J$ and $I$ are the numbers of words in the dialogue context and its response, respectively. Semantic vector space models map each word $x_{j}$ into embeddings $\mathbf{x}_{j}^{emb}$, and $y_{i}$ into $\mathbf{y}_{i}^{emb}$. The semantic representation of a dialogue context $\mathbf{x}$ is then $\mathbf{x}^{emb} = \sum_{j = 1}^{J} w_{j}\mathbf{x}_{j}^{emb}$; for a response $\mathbf{y}$, it is $\mathbf{y}^{emb} = \sum_{i = 1}^{I} v_{i}\mathbf{y}_{i}^{emb}$. Here, $w_{j}$ and $v_{i}$ are importance weights for each word in the sentence.\footnote{We set the importance weights to 0 for a list of stop words (high-frequency words such as articles and prepositions, names, punctuation marks), 1 otherwise.}
The measure of coherence is then defined as the cosine distance of the two semantic vectors of the dialogue context and its response: 
\begin{equation}\label{Eq:1}
\begin{split}
	C\left (\mathbf{x}, \mathbf{y} \right ) &= \cos\left ( \mathbf{x}^{emb}, \mathbf{y}^{emb} \right ) 
\end{split}
\end{equation}


\subsection{Model Overview} \label{SubSec:overview}

End-to-end response generation for dialogue can be formalized as follows:
Given a dialogue context $\mathbf{x}$, a dialogue generator generates the next utterance $\mathbf{y}$. During the training process, the aim for a dialogue generator is to maximize the probability $p\left (\mathbf{y}| \mathbf{x} \right )$ over the training dataset. To encode dialogue contexts that adequately incorporate coherence information, we build our generator based on the cVAE model of \citet{hu2017toward}, which has been used to control text generation with respect to linguistic properties, such as tense or sentiment.

In our model, the response $\mathbf{y}$ is generated conditioned on the previous conversation $\mathbf{x}$, a diversity-promoting latent variable $\mathbf{z}$, and a latent variable $c$ indicating  dialogue coherence; $\mathbf{z}$ and $c$ are independent. The generation probability $p\left (\mathbf{y}|\mathbf{x} \right )$ is defined as:
\begin{equation}\label{Eq:2}
\begin{split}
	p\left (\mathbf{y}|\mathbf{x} \right ) &= \int_{\mathbf{z}, c} p\left (\mathbf{y}|\mathbf{x}, \mathbf{z}, c \right ) p\left (\mathbf{z}, c |\mathbf{x} \right ) \diff\mathbf{z} \diff c \\
    &= \int_{\mathbf{z}, c} p\left (\mathbf{y} | \mathbf{x}, \mathbf{z}, c \right ) p\left (\mathbf{z} | \mathbf{x} \right ) p\left (c | \mathbf{x} \right )\diff\mathbf{z} \diff c
\end{split}
\end{equation}

Unfortunately, optimizing Eq~(\ref{Eq:2}) during training is intractable; therefore, we apply variational inference and optimize instead the variational lower bound:
\begin{equation}\label{Eq:3}
\begin{split}
	\log p\left (\mathbf{y}|\mathbf{x} \right ) = & \log \int_{\mathbf{z}, c} p\left (\mathbf{y} | \mathbf{x}, \mathbf{z}, c \right ) p\left (\mathbf{z}, c | \mathbf{x} \right ) \diff\mathbf{z} \diff c \\
    \geq &\ \mathbb{E}_{q\left ( \mathbf{z} | \mathbf{x}, \mathbf{y} \right )p\left (c | \mathbf{x}, \mathbf{y} \right )}\left [ \log p\left (  \mathbf{y} | \mathbf{x},\mathbf{z},c \right ) \right ] \\
    &- D_{KL}\left ( q\left ( \mathbf{z} | \mathbf{x}, \mathbf{y} \right ) \parallel  p\left (\mathbf{z} | \mathbf{x}  \right ) \right )
\end{split}
\end{equation}
where 
$p\left ( \mathbf{y} | \mathbf{x},\mathbf{z},c \right )$ is the probability of generating utterance $\mathbf{y}$ given $\mathbf{x}$, $\mathbf{z}$ and $c$;
$q\left ( \mathbf{z} | \mathbf{x}, \mathbf{y} \right )$ stands for the approximate posterior distribution of the latent variable $\mathbf{z}$ conditioned on dialogue context $\mathbf{x}$ and the gold response $\mathbf{y}$; 
$p\left (c | \mathbf{x}, \mathbf{y} \right )$ is the measure of coherence between context $\mathbf{x}$ and response $\mathbf{y}$; 
$p\left (\mathbf{z} | \mathbf{x}  \right )$ is the 
true prior distribution of $\mathbf{z}$ conditioned only on dialogue context $\mathbf{x}$;
$D_{KL}\left ( \cdot | \cdot  \right )$ denotes the KL-divergence. 
We assume that both $q\left ( \mathbf{z} | \mathbf{x}, \mathbf{y} \right )$ and $p\left (\mathbf{z} | \mathbf{x}  \right )$ are Gaussian with mean vectors $\mu_{appr}$, $\mu_{true}$ and covariance matrices $\Sigma_{appr}$, $\Sigma_{true}$.

\subsubsection{Model Details}
\begin{figure*}[tb]
	\centering
	\includegraphics[width=0.7\textwidth]{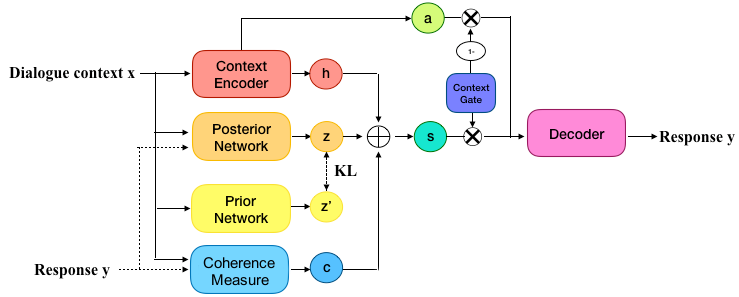}
	\caption{The training process of the generative model. First, the dialogue context is encoded: $\mathbf{h}$ is the final hidden state of the context encoder. Then we derive the diversity-promoting latent variable $\mathbf{z}$. Next, we  compute the latent variable $c$ that corresponds to the measure of coherence between the dialogue context $x$ and the generated response $y$. We concatenate all three vectors into $\mathbf{s}$ to feed the decoder. $\mathbf{a}$ is the attention matrix calculated for every time step of the decoding process.}
    \label{fig:model_training}
\end{figure*}

Optimizing Eq~(\ref{Eq:3}) consists of two parts: (1) minimizing the KL-divergence between the approximate posterior distribution and the true prior distribution of $\mathbf{z}$, (2) maximizing the probability of generating the gold response $\mathbf{y}$ conditioned on dialogue context $\mathbf{x}$ and coherence factors $\mathbf{z}$ and $c$.
Figure \ref{fig:model_training} shows the pipeline of the training procedure. 

\paragraph{Encoder:} First, we encode a dialogue context $\mathbf{x}$ into a hidden state $\mathbf{h}$ using the \emph{context encoder}, which is based on Recurrent Neural Networks (RNNs). Then the \emph{posterior network} encodes both dialogue context $\mathbf{x}$ and gold response $\mathbf{y}$ into a hidden state $\mathbf{h}_{appr}$ followed by two linear transformations $f_{appr}\left ( \cdot  \right )$ and $g_{appr}\left ( \cdot  \right )$ to map $\mathbf{h}_{appr}$ into mean vector $\mu_{appr}$ and covariance matrix $\Sigma_{appr}$. The latent variable $\mathbf{z}$ can be sampled from the distribution $N\left ( \mu_{appr}, \Sigma_{appr} \right )$:
\begin{equation}\label{Eq:4}
\begin{aligned}
\mu_{appr} &= f_{appr}\left ( \mathbf{h}_{appr}  \right ) \\
\Sigma_{appr} &= g_{appr}\left ( \mathbf{h}_{appr}  \right ) \\
q\left (\mathbf{z} | \mathbf{x}, \mathbf{y}  \right ) &=  N\left ( \mu_{appr}, \Sigma_{appr} \right )
\end{aligned}
\end{equation}
The \emph{prior network} in Figure~\ref{fig:model_training} takes a form similar to the posterior network:
\begin{equation}\label{Eq:5}
\begin{aligned}
\mu_{true} & = f_{true}\left ( \mathbf{h}_{true}  \right ) \\
\Sigma_{true} & = g_{true}\left ( \mathbf{h}_{true}  \right ) \\
p\left (\mathbf{z} | \mathbf{x} \right ) & =  N\left ( \mu_{true}, \Sigma_{true} \right )
\end{aligned}
\end{equation}
where $\mathbf{h}_{true}$ is the final hidden state of an RNN encoding only the dialogue context $\mathbf{x}$, and $f_{true}\left ( \cdot  \right )$, $g_{true}\left ( \cdot  \right )$ are linear transformations. 
Code $c$ is given by the \emph{coherence measure} from Eq~(\ref{Eq:1}).

\paragraph{Decoder:} We build an attention-based decoder \cite{bahdanau2014neural,luong2015effective} using RNNs to generate responses conditioned on encoded dialogue context $\mathbf{h}$, diversity signal $\mathbf{z}$, and coherence signal $c$. We concatenate the latent variables $\mathbf{z}$ and $c$ to the context encoder hidden state $\mathbf{h}$ and feed them into the decoder as the initial hidden state $\mathbf{s}_0$, similar to \newcite{hu2017toward}. 

During the decoding process, tokens are generated sequentially under the following probability distribution:
\begin{equation}\label{Eq:6}
\begin{split}
p\left (\mathbf{y} | \mathbf{x}, \mathbf{z}, c \right ) &= \prod_{i=1}^{I} p\left (y_{i}| y^{< i}, \mathbf{x}, \mathbf{z}, c \right ) \\
&= \prod_{i=1}^{I} g\left (y_{i-1}, \mathbf{s}_i, \mathbf{a}_i \right )
\end{split}
\end{equation}
where $I$ is the length of the produced response; $g\left ( \cdot  \right )$ is an RNN; $\mathbf{s}_i$ is the hidden state of the decoder at time step $i$ which is conditioned on the previously generated token $y_{i-1}$, the previous hidden state $\mathbf{s}_{i-1}$, and the weighted attention vector $\mathbf{a}_i$:
\begin{align}
\label{Eq:8}
\mathbf{s}_i & = f\left ( y_{i-1}, \mathbf{s}_{i-1}, \mathbf{a}_i  \right ) \\
\label{Eq:7}
\mathbf{a}_i & = \sum_{j=1}^{J}\mathbf{w}_{ij}\mathbf{h}_i
\end{align}
where $J$ is the number of tokens of the dialogue context; $\mathbf{h}_i$ is the $i^{th}$ hidden state of the encoder; the attention weight $\mathbf{w}_{ij}$ of each context hidden state $\mathbf{h}_i$ is computed following \newcite{luong2015effective}.

\paragraph{Context Gate:} To increase the influence of code $c$, we introduce the context gate $k$. Unlike \newcite{tu2016context}, whose context gate assigns an element-wise weight to the input signal deriving from the encoder RNN, we build the context gate conditioned only on the coherence signal:
\begin{equation}\label{Eq:9}
k_i = \lambda \sigma \left ( c-{c}'_i \right )
\end{equation}
where $\sigma$ is the sigmoid function; $\lambda$ is a bias term;\footnote{We set $\lambda$ empirically against the development set.} $c$ is the target value of the measure of coherence, calculated by $C\left ( \mathbf{x}, \mathbf{y} \right)$ (see Section~\ref{SubSec:coherence_def}); ${c}'_i$ is the measure of coherence between the dialogue context and the generated prefix sentence at time step $i$, calculated by $C\left ( \mathbf{x}, \mathbf{y}^{<i} \right )$. Now Eq~(\ref{Eq:8}) with the context gate applied to $\mathbf{s}_{i}$ can be rewritten as:
\begin{equation}\label{Eq:10}
\mathbf{s}_i = f\bigl( y_{i-1}, \left (1 - k_i \right ) \circ \mathbf{s}_{i-1}, k_i \circ \mathbf{a}_i  \bigr)
\end{equation}
where $\circ$ denotes element-wise multiplication.

The coherence-informed context gate aims to dynamically control the ratio at which preceding dialogue context and previously generated tokens of the current response contribute to the generation of the next token in the response.

\subsection{Training} \label{SubSec:generator}

Our generator is trained similarly to \newcite{hu2017toward}. The objective function is a weighted combination of three losses (generation, coherence, and diversity):
\begin{equation}\label{Eq:11}
\begin{split}
\mathcal{L} = \mathcal{L}_{G} + \lambda_c\mathcal{L}_{c} + \lambda_\mathbf{z}\mathcal{L}_{\mathbf{z}} 
\end{split}
\end{equation}
To teach the generator to produce responses close to the training data, we maximize the generation probability of the training response $\log p\left ( \mathbf{y} | \mathbf{x}  \right )$ given the dialogue context according to Eq~(\ref{Eq:2}). During training, we set $\mathcal{L}_{G} = -\log p\left ( \mathbf{y} | \mathbf{x}  \right )$ and minimize the following:
\begin{equation}\label{Eq:12}
\begin{split}
\mathcal{L}_{G} = & D_{KL}\left ( q\left ( \mathbf{z} | \mathbf{x}, \mathbf{y} \right ) \parallel  p\left (\mathbf{z} | \mathbf{x}  \right ) \right ) \\
& - \mathbb{E}_{q\left ( \mathbf{z} | \mathbf{x}, \mathbf{y} \right )p\left (c | \mathbf{x}, \mathbf{y} \right )}\left [ \log p\left (  \mathbf{y} | \mathbf{x},\mathbf{z},c \right ) \right ]
\end{split}
\end{equation}
Apart from the generation loss, the coherence measure provides an extra learning signal $\mathcal{L}_{c}$ which pushes the generator to produce responses that match the coherence signal given by the latent variable $c$.
\begin{equation}\label{Eq:13}
\begin{split}
\mathcal{L}_{c} = -\mathbb{E}_{p\left ( \mathbf{z} |  \mathbf{x} \right )p\left ( c \right )}\left [ \log p\left ( c | \mathbf{x}, \widetilde{G}\left ( \mathbf{x}, \mathbf{z}, c \right ) \right ) \right ]
\end{split}
\end{equation}
In Eq~(\ref{Eq:13}), $p\left ( c \right )= N\left ( 0, 1 \right )$ is the prior distribution of the coherence variable $c$. 
To ensure that the loss is differentiable, we cannot sample words from the response vocabulary. Instead we define $\widetilde{G}\left ( \mathbf{x}, \mathbf{z}, c \right ) = \mathbf{y}^s = \left \{ \mathbf{y}_1^s, \dots \mathbf{y}_i^s, \dots \mathbf{y}_I^s  \right \}$ as the sequence of output word probability distributions. 
$p\left ( c | \mathbf{x}, \widetilde{G}\left ( \mathbf{x}, \mathbf{z}, c \right ) \right )$ is predicted by the coherence measure defined in Eq~(\ref{Eq:1}) with $\mathbf{y}^{emb}$ set as:
\begin{equation}\label{Eq:15}
\mathbf{y}^{emb} = \sum_{i = 1}^{I}  {\mathbf{y}^s_j}^\top M_{glv}
\end{equation}
where $M_{glv}$ is the word embedding matrix trained using GloVe (Section~\ref{SubSec:coherence_def}).

The last component in Eq~(\ref{Eq:11}) is the independent constraint $\mathcal{L}_{\mathbf{z}}$ that forces the soft distribution over the generated response $\widetilde{G}$ to be diverse, so that it is able to faithfully reproduce the latent variable $\mathbf{z}$:
\begin{equation}\label{Eq:16}
\mathcal{L}_{\mathbf{z}} = -\mathbb{E}_{p\left ( \mathbf{z} | \mathbf{x} \right )p\left ( c \right )}\left [ \log q\left ( \mathbf{z} | \mathbf{x}, \widetilde{G}\left ( \mathbf{x}, \mathbf{z}, c \right ) \right ) \right ]
\end{equation}
where $q\left ( \mathbf{z} | \mathbf{x}, \widetilde{G}\left ( \mathbf{x}, \mathbf{z}, c \right ) \right )$ is predicted by the posterior network with ${\mathbf{y}^s_j}$ as the soft input to the RNN encoder at each time step $j$.


\subsection{Inference} \label{SubSec:testing}

\begin{figure*}[tb]
	\centering
	\includegraphics[width=0.7\textwidth]{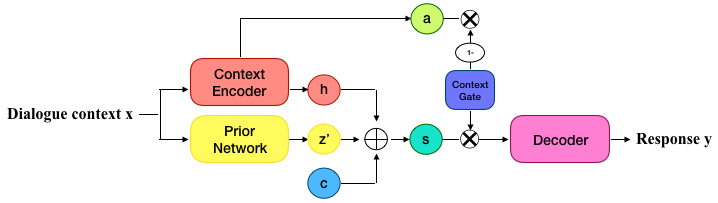}
	\caption{The inference process of the generative model, where the latent variable $c$ is given as an input.
    \label{fig:model_testing}} 
\end{figure*}

Figure \ref{fig:model_testing} shows the inference process of the generative model. Given a dialogue context $\mathbf{x}$ and an expected coherence value $c$, the context encoder first encodes the dialogue context into a hidden state $\mathbf{h}$. The prior network then generates a sample $\mathbf{z'}$ conditioned on the dialogue context. The decoder is initialized with $\mathbf{s}$, i.e., the concatenation of $\mathbf{h}$, $\mathbf{z}$ and $c$. During decoding, the next word is generated via the context gate modulating between the attention-reweighted context and the previously generated words of the response. 

\section{Dataset and Filtering}\label{sec:dataset}

\subsection*{Dataset for Generator}

We train and evaluate our models on the OpenSubtitles corpus \cite{lison_opensubtitles2016:_2016} with automatic dialogue turn segmentation \cite{lison_automatic_2016}.\footnote{\url{http://www.opensubtitles.org}} A training pair consists of a dialogue context and a corresponding response. We consider three consecutive turns as the dialogue context and the following turn as the response. From a total of 65M instances, we select those that have context and response lengths of less than 120 and 30 words, respectively. We create two datasets:
\begin{enumerate}[itemsep=2pt,leftmargin=12pt]
\item \textbf{OST} (plain OpenSubtitles) consists of 2M/4K/4K instances as our training/development/test sets, selected randomly from the whole corpus;
\item \textbf{fOST} (filtered OpenSubtitles) contains the same amount of instances, but randomly selected only among those that have a measure of coherence score 
 $C\left(\mathbf{x}, \mathbf{y}\right) \geq 0.68$.\footnote{The coherence score is calculated as shown in Eq~(\ref{Eq:1}). We observed that the scores on the training set follow a normal distribution with a slight tail on the negatively correlated side, so we fit a normal distribution to the data with parameters $N(0.25, 0.22)$ and set the cut-off to $+2\sigma$. A histogram of coherence scores is shown in Figure \ref{fig:histogram} in Supplemental Material~\ref{sec:coherence-cutoff}.}
\end{enumerate}

Filtering of the OpenSubtitles corpus is motivated by the fact that by removing the video and audio modalities which the subtitles originally accompanied, we are very often left with incomplete and incoherent dialogues.
Therefore, by keeping dialogues with high coherence scores, we aim at building a high quality corpus with (1) more semantically coherent and topically related contexts and responses, and (2) fewer general and dull responses. Table~\ref{table:dataset} shows the coherence and diversity metrics (cf.~Section~\ref{subsec:eval}) between OST and fOST. Unsurprisingly, coherence for fOST is much higher than OST, with a slightly higher diversity. We list dialogue examples for different coherence scores in Supplemental Material~\ref{sec:examples-coherence}.

\begin{table}[t!]\label{table:dataset}
\centering
\small
\begin{tabular}{lcccc} 
  \hline
  \bf Dataset & \bf Coh & \bf D-1\% & \bf D-2\% & \bf D-Sent\%\\
  \hline
  OST & 0.390 & 14.3 & 57.9 & 83.8 \\
  fOST & 0.801 & 15.5 & 62.9 & 89.3 \\
  \hline
\end{tabular}
\caption{Coherence and diversity metrics\footnotemark{} for the OST and fOST datasets (see Section~\ref{sec:dataset} for the datasets and Section~\ref{subsec:eval} for metrics definition).}
\end{table}
\footnotetext{Note that Distinct-1 and Distinct-2 are computed on a randomly selected subsets of 4k responses.}

\subsection*{Dataset for Coherence Measure}
In order to accurately measure coherence on our domain using the semantic distance as defined in Section~\ref{SubSec:coherence_def}, we train GloVe embeddings on the full OpenSubtitles corpus (i.e.\ 100K movies).


\section{Experiments}\label{sec:experiments}

Our generator model, ablative variants, and baselines are implemented using the publicly available OpenNMT-py framework \cite{opennmt} based on \newcite{bahdanau2014neural} and \newcite{luong2015effective}.
We used the publicly available glove-python package\footnote{\url{https://github.com/maciejkula/glove-python}} to implement our coherence measure.

We experiment on two versions of our model: (1) cVAE with the coherence context gate as described in Section~\ref{SubSec:generator} (\emph{cVAE-XGate}), (2) cVAE with the original context gate implementation of \cite{tu2016context} (\emph{cVAE-CGate}). For each of these, we consider
the main variant where the input coherence measure $c$ is preset to a fixed ideal value as estimated on development data (1.0 for OST and 0.95 for fOST), as well as an oracle variant where we use the true coherence measure between the context and the gold-standard response in the test set (indicated with ``(C)'' in Tables~\ref{Tab:test_ost} and~\ref{Tab:test_fost}).

We compare against two baseline models: (1) a vanilla E-D with attention (\emph{Attention}) \cite{luong2015effective}; (2) an enhancement where output beams are rescored using the maximum mutual information anti-language model (MMI-antiLM) of \citet{JiweiLi:diversity2015} (\emph{MMI}).

\subsection{Parameter Settings}\label{subsec:parameters}

We set our model parameters based on preliminary experiments on the development data.

We use 2-layer RNNs with LSTM cells \cite{hochreiter_long_1997} with input/hidden dimension of 128 for both the context encoder and the decoder. The dropout rate is set to 0.2 and the Adam optimizer \cite{kingma2014adam} is used to update the parameters. A vocabulary of 25,000 words is shared between the encoder and the decoder.

Both the posterior network and prior network for the latent variable learning are built with 2-layer LSTM RNNs with input/hidden dimension of 64. The dimension of the latent variable $\mathbf{z}$ is set to 20. Same as for the encoder and decoder, the dropout rate is 0.2 and the Adam optimizer is used to update the parameters.

The window size for GloVe computation in our coherence measure is set to 10.


\subsection{Evaluation metrics}\label{subsec:eval}

We use a number of metrics to evaluate the outputs of our models:
\begin{itemize}[nosep,leftmargin=12pt]
\item \textbf{BLEU, B1, B2, B3} -- the word-overlap score against gold-standard responses \cite{papineni2002bleu} used by the vast majority of recent dialogue generation works \cite{zhao2017learning, yao2017towards, li2017learning, li2016simple,sordoni:2015,JiweiLi:diversity2015, MS2017knowledge}. BLEU in this paper refers to the default BLEU-4, but we also report on lower $n$-gram scores (B1, B2, B3).\footnote{We use the Multi-BLEU script from OpenNMT to measure BLEU scores.}

\item \textbf{Coh} -- our novel GloVe-based coherence score calculated using Eq~(\ref{Eq:1}) showing the semantic distance of dialogue contexts and generated responses.

\item \textbf{D-1, D-2, D-Sent} -- common metrics used to evaluate the diversity of generated responses \cite[e.g.][]{JiweiLi:diversity2015,xu2017neural, xing2017topic, dhingra2017towards}: the proportion of distinct unigrams, bigrams, and sentences in the outputs.
\end{itemize}


\section{Results}\label{sec:results}

\begin{table*}[tb] 
\centering
\small
\begin{tabular}{l|l|cccccccc}
  \hline
  \bf Training data & \bf Model & \bf BLEU\% &\bf B1\% & \bf B2\% & \bf B3\% & \bf Coh & \bf D-1\% &  \bf D-2\% & \bf D-Sent\% \\
  \hline 
  \hline
   \multirow{6}{*}{OST} & Attention     & 1.32  & 10.92 & 3.85 & 2.10 & 0.293 & \phantom{0}3.4 & 14.2 & 25.6 \\
   & MMI              & 1.31  & 11.06 & 3.88 & 2.09 & 0.284 & \phantom{0}3.3 & 14.6 & 28.2 \\
   & cVAE-CGate (C)   & 1.58  & 11.86 & 4.45 & 2.48 & 0.311 & \phantom{0}4.1 & 15.0 & 28.2 \\
   & cVAE-XGate (C)   & 1.51  & 13.38 & 4.97 & 2.58 & 0.324 & \phantom{0}3.9 & 14.5 & 29.8 \\
   & cVAE-CGate (1.0) & \bf{1.60}  & \bf{17.08} & \bf{5.78} & \bf{2.86} & 0.404 & \bf{\phantom{0}5.0} & \bf{27.1} & 79.7 \\
   & cVAE-XGate (1.0) & 1.44  & 15.83 & 5.34 & 2.62 & \bf{0.41}3 & \phantom{0}4.5 & 22.6 & \bf{80.2} \\
  \hline
  \multirow{6}{*}{fOST} & Attention     & 1.79  & 15.43 & 5.65 & 2.94 & 0.758 & \bf{11.9} & 41.8 & 92.7 \\
   & MMI              & 1.99  & 16.24 & \bf{6.06} & 3.22 & 0.764 & \bf{11.9} & 44.5 & 95.8 \\
   & cVAE-CGate (C)   & \bf{2.10}  & 15.98 & 6.05 & \bf{3.35} & 0.728 & \bf{11.9} & 37.6 & 88.4 \\
   & cVAE-XGate (C)   & 1.85  & \bf{16.44} & 5.94 & 3.07 & 0.706 & 10.3 & 31.2 & 80.4 \\
   & cVAE-CGate (0.95)& 2.02  & 15.52 & 5.78 & 3.16 & \bf{0.767} & 10.6 & \bf{44.8} & \bf{98.7} \\
   & cVAE-XGate (0.95)& 1.64  & 14.43 & 5.20 & 2.70 & 0.745 & \phantom{0}9.0 & 36.9 & \bf{98.7} \\
  \hline
\end{tabular}
\caption{Evaluation results on the \underline{OST} test set (see Section~\ref{sec:experiments} for model description and Section~\ref{subsec:eval} for metrics definition). Note that the cVAE-CGate(C) / cVAE-XGate(C) models use the true $c$ value between the context and the gold response as input. Other cVAE-CGate / cVAE-XGate models use fixed values for $c$ selected on dev sets shown in brackets. 
BLEU score reported here is BLEU-4; B1, B2 and B3 denote lower $n$-gram BLEU scores.
\label{Tab:test_ost}}
\end{table*}

\begin{table*}[tb] 
\centering
\small
\begin{tabular}{l|l|cccccccc} 
  \hline
  \bf Training data & \bf Model & \bf BLEU\% &\bf B1\% & \bf B2\% & \bf B3\% & \bf Coh & \bf D-1\% &  \bf D-2\% & \bf D-Sent\% \\
  \hline 
  \hline
   \multirow{6}{*}{OST} & Attention & 0.86  & \phantom{0}8.34 & \phantom{0}2.79 & 1.45 & 0.284 & \phantom{0}3.6 & 14.6 & 29.4 \\
   & MMI                 & 0.89  & \phantom{0}8.47 & \phantom{0}2.89 & 1.48 & 0.278 & \phantom{0}3.7 & 15.3 & 31.5 \\
   & cVAE-CGate (C)      & 1.64  & 10.20 & \phantom{0}4.17 & 2.40 & 0.329 & \phantom{0}5.1 & 19.4 & 35.8 \\
   & cVAE-XGate (C)      & 1.80  & 11.70 & \phantom{0}4.90 & 2.83 & 0.359 & \phantom{0}5.2 & 19.2 & 39.7 \\
   & cVAE-CGate (1.0)    & 2.25  & 16.82 & \phantom{0}6.81 & 3.70 & 0.422 & \bf{\phantom{0}5.4} & \bf{28.2} & 81.0 \\
   & cVAE-XGate (1.0)    & \bf{2.41}  & \bf{18.62} & \phantom{0}\bf{7.56} & \bf{4.09} & \bf{0.434} & \phantom{0}4.8 & 23.4 & \bf{84.0} \\
  \hline
   \multirow{6}{*}{fOST} & Attention & 3.84  & 16.65 & \phantom{0}8.72 & 5.54 & 0.803 & \bf{12.8} & 43.4 & 88.7 \\
   & MMI                 & 3.84  & 16.81 & \phantom{0}8.78 & 5.57 & 0.803 & 12.6 & 42.5 & 88.8 \\
   & cVAE-CGate (C)      & 4.58  & 17.64 & \phantom{0}9.53 & 6.30 & 0.796 & 12.4 & 41.6 & 85.5 \\
   & cVAE-XGate (C)      & 4.33  & 18.43 & \phantom{0}9.59 & 6.11 & 0.783 & 10.7 & 33.1 & 78.8 \\
   & cVAE-CGate (0.95)   & \bf{4.98}  & 20.95 & \bf{10.93} & \bf{7.02} & \bf{0.814} & 12.1 & \bf{51.4} & \bf{98.2} \\
   & cVAE-XGate (0.95)   & 4.47  & \bf{20.98} & 10.43 & 6.50 & 0.797 & 10.4 & 42.5 & 97.6 \\
  \hline
\end{tabular}
\caption{Evaluation results on the \underline{fOST} test set (see Section~\ref{sec:experiments} and Table~\ref{Tab:test_ost} for model description; see Section~\ref{subsec:eval} for metrics definition). 
BLEU score reported here is BLEU-4; B1, B2 and B3 denote lower $n$-gram BLEU scores.
\label{Tab:test_fost}}
\end{table*}

All model variants described in Section~\ref{sec:experiments} are trained on both OST and fOST datasets.
Tables~\ref{Tab:test_ost} and~\ref{Tab:test_fost} present the scores of all models tested on the OST  and fOST test sets, respectively. 
Note that in addition to testing the models on the respective test sections of their training datasets, we also test them on the other dataset (OST-trained models on fOST and vice-versa).
This way, we can observe the performance of the fOST-trained models in more noisy contexts and see how good the OST-trained models are when evaluated against coherent responses only.

Given all the evaluated model variants, we can observe the effects and contributions of the individual components of our setup:

\begin{itemize}[itemsep=2pt,leftmargin=12pt]
\item \emph{Data filtering:} The models trained on fOST consistently outperform the same models trained on OST -- for all evaluation metrics and on both test sets. This shows that coherence-based training data filtering is generally beneficial.
 
\item \emph{cVAE-Context Gate models:} Nearly all cVAE-based models perform markedly better than the baselines w.r.t.\ BLEU, coherence, and diversity.\footnote{ 
We performed paired bootstrap re-sampling for the best cVAE model and the best baseline model in each experiments set (Table \ref{Tab:test_ost} and Table \ref{Tab:test_fost}) as is done for MT \citep{koehn2004statistical}, which confirmed statistical significance at 99\% confidence level for all cases except for models trained on fOST and tested on OST (bottom half of Table~\ref{Tab:test_ost}).}

If we look at models trained on OST and tested on fOST (the top half of Table~\ref{Tab:test_fost}), we can see that all cVAE-based models, especially cVAE-XGate, are able to learn to produce coherent and diverse response even when trained on a noisy, incoherent corpus. Examples of responses generated by the baseline MMI model and by cVAE-XGate in Figure~\ref{fig:show_case} show that cVAE-XGate mostly produces more diverse and coherent responses than MMI.

\item \emph{Preset $c$ vs.\ oracle models with gold-standard $c$:} 
Table~\ref{Tab:test_ost} shows that on the noisy OST test set, cVAE-based models using the gold-standard value of $c$ achieve higher BLEU scores than models using preset $c$.
This is expected since many gold-standard responses in the unfiltered set have a low coherence score; -- the model can generate a more generic response if the gold-standard $c$ is low.
The models with preset $c$ always attempt to generate coherent responses, which is apparent from the other metrics: Coh and D-Sent are consistently higher than for models using gold-standard $c$.

On the fOST test set where only high-coherence responses are expected, models using fixed $c$ consistently reach higher scores in all metrics including BLEU  (see Table~\ref{Tab:test_fost}).
This shows that in general, using a preset constant value of $c$ works well, even better than using the gold-standard $c$.
\end{itemize}

In sum, using our coherence measure both for data filtering and inside the models leads to output performance improvements.

\begin{figure*}[tb]
\begin{center}
\small
\begin{tabular}{|c|L{0.5cm} L{6.5cm}| L{7cm}|}\hline
 & & {\bf Dialogue context} & {\bf Response}  \\\hline\hline
 & A: & i have an audition at 4:00. & B-GT: you volunteered.  \\ 
1& B: & apparently now i'm \uline{cooking dinner}. & B-MMI: i don 't know. \\ 
 & A: & how did this all happen? & B-XGT: well , i'm going to \uline{have a dinner}. \\\hline\hline

 & A: & it's not that easy. & B-GT: the waiting list would take too long.  \\ 
2& B: & we can't contact potential donors. & B-MMI: i don't know. \\ 
 & A: & if we want to \uline{save frank}, his brother is his only hope & B-XGT: we can't \uline{save him}. \\\hline\hline

 & A: & that \uline{fish} ate my \uline{fish}! & B-GT: you'll wait till next year to win one of these.  \\ 
3& B: & \uline{you cost me} \pounds 30,000 & B-MMI: i'm sorry. \\ 
 & A: & sorry, folks. & B-XGT: \uline{you owe me} a little \uline{fish}.\\\hline\hline

 & A: & was it what you wanted? & B-GT: you like model planes?  \\ 
4& B: & no. & B-MMI: i wanted to know.\\ 
 & A: & i wanted an \uline{airplane}. & B-XGT: i wanted a \uline{helicopter}.\\\hline\hline

 & A: & great. & B-GT: yeah, a vanilla decaf latte. \\ 
5& B: & thanks so much. & B-MMI: no. \\ 
 & A: & \uline{vanilla decaf latte?} & B-XGT: \uline{yeah} , that 's a \uline{good coffee} \\\hline\hline

 & A: & to make everybody happy. & B-GT: and stick it in your mouth. \\ 
6& B: & which means that sometimes & B-MMI: and you got to take a shower. \\ 
 & A: & you got to take a pile of poop & B-XGT: and you got to take a walk. \\\hline\hline

 & A: & now you'll knock again and say you forgot to ask & B-GT: should i knock and make her come here for that? \\ 
7& B: & if she has a friend for your friend. & B-MMI: you're \uline{crazy}! \\ 
 & A: & are you \uline{crazy}? & B-XGT: i'm \uline{crazy}, i'm \uline{crazy}. \\\hline\hline

 & A: & i knew you would say that. & B-GT: you are all way too pretty to be arguing. \\ 
8& B: & mr. kerem is a \uline{charming and decent guy}. & B-MMI: he's a \uline{very good man}. \\ 
 & A: & and he's really into you. & B-XGT: i didn't say that. \\\hline
\end{tabular}
\end{center}

\caption{
Generation output examples comparing two models trained on the OST dataset and tested on the fOST test set: the MMI baseline (B-MMI) and our cVAE-XGate (1.0) model (B-XGT). B-GT is the ground-truth response from the test set. 
The three sequential dialog turns on the left are the preceding 
dialogue context used to generate the responses.
Corresponding topical phrases are underlined. We can see that cVAE-XGate (1.0) mostly produces markedly more coherent and specific outputs than MMI (1-5). In some cases, it is comparable with MMI (6-7) and occasionally, it is less coherent (8).}
\label{fig:show_case}
\end{figure*}

\section{Related Work}\label{sec:related}

Our work fits into the context of the very active area of end-to-end generative conversation models, where neural E-D approaches have been first applied by \citet{Vinyals:2015} and extended by many others since.

Many works address the lack of diversity and coherence in E-D outputs \cite{sountsov_length_2016,wei_why_2017} but do not attempt to model coherence directly, unlike our work:
\citet{JiweiLi:diversity2015} use anti-LM reranking; \citet{li2016simple} modify the beam search decoding algorithm, similar to \citet{shao_generating_2017} in addition to using a self-attention model.  \citet{mou_sequence_2016} predict keywords for the output in a preprocessing step while \citet{wu_neural_2018} preselect a vocabulary subset to be used for decoding.
\citet{li_persona-based_2016} focus specifically on personality generation (using personality embeddings) and \citet{wang_steering_2017} promote topic-specific outputs by language-model rescoring and sampling.

A lot of recent works explore the use of additional training signals and VAE setups in dialogue generation. In contrast to this paper, they do not focus explicitly on coherence:
\citet{asghar_deep_2017} use reinforcement learning with human-provided feedback, 
\citet{li2017learning} use a RL scenario with length as reward signal. \citet{li_adversarial_2017} add an adversarial discriminator to provide RL rewards (discriminating between human and machine outputs), \citet{xu2017neural} use a full adversarial training setup.
The most recent works explore the usage of VAEs:
\citet{cao_latent_2017} explore a vanilla VAE setup conditioned on dual encoder (for contexts and responses) during training, the model of \citet{Serban2017AHL} uses a VAE in a hierarchical E-D model.
\citet{shen_conditional_2017} use a cVAE conditioned on sentiment and response genericity (based on a handwritten list of phrases). \citet{shen_improving_2018} combine a cVAE with a plain VAE in an adversarial fashion.

We also draw on ideas from other areas than dialogue generation to build our models: \newcite{tu2016context}'s context gates originate from machine translation and \newcite{hu2017toward}'s cVAE training stems from free-text generation.

\section{Conclusions and Future Work}\label{sec:concl}

We showed that explicitly modeling coherence and optimizing towards coherence and diversity leads to better-quality outputs in dialogue response generation.
We introduced three extensions to current encoder-decoder response generation models:
(1) we defined a measure of coherence based on GloVe embeddings \cite{pennington2014glove}, 
(2) we filtered the OpenSubtitles training corpus \cite{lison_automatic_2016} based on this measure to obtain coherent and diverse training instances,
(3) we trained a cVAE model based on \cite{hu2017toward} and \cite{tu2016context} that uses our coherence measure as one of the training signals.
Our experimental results showed a considerable improvement in the output quality over competitive models, which demonstrates the effectiveness of our approach.

In future work, we plan to replace the GloVe-based measure of coherence with a trained discriminator that distinguishes between coherent and incoherent responses \cite{li2016neural}. This will allow us to use extend the notion of coherence to account for phenomena such as topic shifts. We also plan to verify the results with a human evaluation study.


\section*{Acknowledgements}

This research received funding from the EPSRC project MaDrIgAL (EP/N017536/1). The Titan Xp used for this research was donated by the NVIDIA Corporation.


\bibliography{emnlp2018}
\bibliographystyle{acl_natbib_nourl}


\clearpage
\onecolumn
\appendix
\begin{center}
\Large\bf
\thetitle: Supplemental Material
\end{center}

\section{Determining Cut-off Coherence Score}
\label{sec:coherence-cutoff}

\begin{figure}[h]
	\centering
	\includegraphics[width=0.47\textwidth]{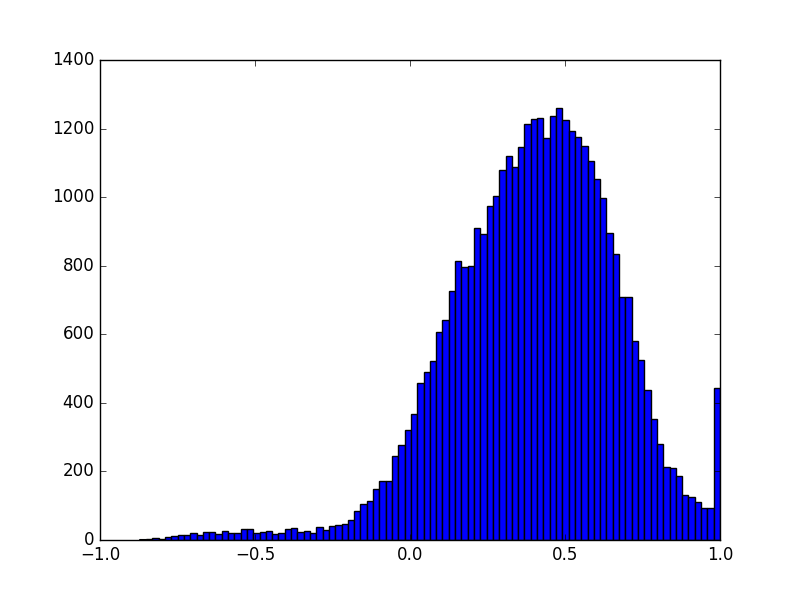}
	\caption{Histogram of coherence scores on the training set before filtering.}
    \label{fig:histogram}
\end{figure}

As shown in Figure~\ref{fig:histogram}, the scores on the training set roughly follow a normal distribution with a slight tail on the negatively correlated side. We make the assumption that the data fit a normal distribution and estimate parameters $\mu=0.25$ and $\sigma=0.22$. We set the cut-off to $+2\sigma$ so that it accounts for 95\% of the scores and does not severely filter the number of resulting examples in the dataset.

\section{Dialogue Examples from the Training Set}
\label{sec:examples-coherence}

Tables~\ref{table:coherence_1}-\ref{table:coherence_4} show dialogue examples from the training set with different coherence scores. The instances in Table~\ref{table:coherence_1} are strongly lexically coherent by repeating the same words in the response when coherence scores are over 0.9. When scores are between 0.7 and 0.9 (Table~\ref{table:coherence_2}), the responses use synonyms to words in the context but still remain semantically coherent. When scores are between 0.4 and 0.6 (Table~\ref{table:coherence_3}), the responses are coherent but very dull; they are getting incoherent when scores are lower (Table~\ref{table:coherence_4}).

\begin{table*}[p]
\centering
\small
\begin{tabular}{lll} 
  \hline
  Dialogue contexts & Responses \\
  \hline
  yeah? \#\# you're new in town, right? & you call this a town? \\
  so you can have marie all to yourself? \#\# you're so selfish, catherine. & you've always been selfish. \\
  wait. \#\# where are you going? \#\#  to find the president. & the president, he lives! \\
  that's not good! \#\# it 's fine. it itches a bit at first, but then it stops. & of course it itches! \\
  you stay out of this. \#\# so you want water, huh? \#\# that's right. & well, i got water. \\
  \hline
\end{tabular}
\caption{Dialogue examples with coherence score $C(\mathbf{x}, \mathbf{y})>0.9$ (``\#\#'' in the context denotes turn boundaries).}\label{table:coherence_1}
\end{table*}

\begin{table*}[p]
\centering
\small
\begin{tabular}{L{10cm}L{5cm}} 
  \hline
  Dialogue contexts & Responses \\
  \hline
  not quite yet. \#\# call your grandfather to pick me up. \#\# i want to go home. & grandpa 's not here . \\
  some kind of whisky nobody's ever heard of. \#\# why don't you bring your own bottle? & give him the best bourbon you got, hot stuff, and don't be gone too long. \\
  put your head on my shoulder? \#\# denny \#\# i just want to remember. & i don't think my neck even bends, anymore. \\
  i don't even know where to start. \#\# kitchen. \#\# definitely the kitchen. & specifically the stove. \\
  the problem is, the liquid just stays in your gut. \#\# i don't know what to do. & well, obviously it's not getting absorbed into the bloodstream.\\
  \hline
\end{tabular}
\caption{Dialogue examples with coherence score $0.7<C(\mathbf{x})<0.9$ (``\#\#'' in the context denotes turn boundaries).}\label{table:coherence_2}
\end{table*}

\begin{table*}[p]
\centering
\small
\begin{tabular}{L{10cm}L{5cm}} 
  \hline
  Dialogue contexts & Responses \\
  \hline
  i'm gonna hold it. \#\# take a look at it. \#\# make sure it was an accident. & doesn't sound right. \\
  in fact, thank you for underplaying it. \#\# so the boy becomes a man & it's amazing. \\
  oh, sting? \#\# little bit. \#\# can we stop at the drug store? & oh, uh, don't worry. \\
  there's no one in there! \#\# it's not gonna happen. \#\# it's a private party. & nice going .\\
  give me that pad. \#\# what are you gonna do? \#\# just watch the old mastermind. & what are you doing? \\
  \hline
\end{tabular}
\caption{Dialogue examples with coherence score $0.4<C(\mathbf{x}, \mathbf{y})<0.6$ (``\#\#'' in the context denotes turn boundaries).}\label{table:coherence_3}
\end{table*}

\begin{table*}[p]
\centering
\small
\begin{tabular}{L{10cm}L{5cm}} 
  \hline
  Dialogue contexts & Responses \\
  \hline
   otherwise it's all over do whatever he wants \#\# alright. \#\# so? & how much do you want? \\
  i'll let it go. \#\# don 't worry. \#\# i've totally let it go. & it's all water under the bridge. \\
  i'm so sorry. \#\# mag? \#\# thank you. & i'm going to go for a walk. \\
  i dont want it. \#\# why \#\# ok & you want the car.\\
  he's thinking about leaving the theatre. \#\# we've saved you a seat here. & come on. look at her hair. \\
  \hline
\end{tabular}
\caption{Dialogue examples with coherence score $C(\mathbf{x}, \mathbf{y})<0.4$ (``\#\#'' in the context denotes turn boundaries).}\label{table:coherence_4}
\end{table*}

\end{document}